\documentclass{article}

\usepackage{PRIMEarxiv}
\usepackage[utf8]{inputenc} % allow utf-8 input
\usepackage[T1]{fontenc}    % use 8-bit T1 fonts
\usepackage{hyperref}       % hyperlinks
\usepackage{url}            % simple URL typesetting
\usepackage{booktabs}       % professional-quality tables
\usepackage{amsfonts}       % blackboard math symbols
\usepackage{nicefrac}       % compact symbols for 1/2, etc.
\usepackage{microtype}      % microtypography
\usepackage{lipsum}
\usepackage{fancyhdr}       % header
\usepackage{graphicx}       % graphics
\usepackage{listings}
\usepackage{mdframed}
\usepackage{xcolor}
\usepackage[round]{natbib}
\usepackage{comment}
\usepackage{tabularx}
\usepackage{float}
\graphicspath{{img/}}     % organize your images and other figures under media/ folder

\newfloat{listing}{htbp}{loa}  
\floatname{listing}{Listing} 

\title{GPT-4 as an Agronomist Assistant? Answering Agriculture Questions Using Large Language Models}

\author{
  Bruno Silva, Leonardo Nunes, Roberto Estevão, Vijay Aski, Ranveer Chandra \\
  Microsoft Research \\
 \texttt{\{brunosilva, lnunes,  robertode, vaski, ranveer\}@microsoft.com} \\
}

\begin{document}
\maketitle

\begin{abstract} 

Large language models (LLMs) have demonstrated remarkable capabilities in
natural language understanding and generation across various domains, including
healthcare and finance.  For some tasks that require
intelligence, LLMs achieve similar or better performance than trained human
beings, therefore it is reasonable to employ human exams (e.g., certification
tests) to assess the performance of language models.  In this study, we present
a comprehensive evaluation of popular LLMs, such as Llama 2, GPT-3.5 and GPT-4, on
their ability to answer agriculture-related questions.  In our evaluation, we
also employ RAG (Retrieval-Augmented Generation) and ER (Ensemble Refinement) techniques, which combine
information retrieval, generation capabilities, and prompting strategies to improve the models'
performance.  To demonstrate the capabilities of LLMs, we selected agriculture
exams and benchmark datasets from three of the largest agriculture producer
countries: Brazil, India, and the USA.  Our analysis highlights
GPT-4's ability to achieve a passing score on exams to earn credits for renewing
agronomist certifications, answering 93\% of the questions correctly and
outperforming earlier general-purpose models (GPT-3.5), which achieved 88\%
accuracy. On one of our evaluation datasets that had published student scores, GPT-4 obtained the highest
performance when compared to human subjects. 
This performance suggests that GPT-4 could potentially pass on major
graduate education admission tests or even earn credits for renewing agronomy
certificates. We also explore the models' capacity to address general
agriculture-related questions and generate crop management guidelines for
Brazilian and Indian farmers, utilizing robust datasets from the Brazilian Agency of
Agriculture (Embrapa) and graduate program exams from India.  The results suggest that GPT-4, 
ER, and RAG can contribute meaningfully to agricultural education, assessment, and crop management
practice, offering valuable insights to farmers and agricultural professionals
from Brazil, India, and the USA.  Implications of these findings are discussed
in terms of the potential uses of GPT-4, ER, and RAG in agricultural education,
assessment, and practice. We emphasize the importance of addressing challenges
related to accuracy and safety while harnessing the power of LLMs effectively
and responsibly in the agriculture domain.

\end{abstract}

\keywords{GPT-4 \and Agriculture \and Retrieval Augmented Generation \and Prompt Engineering \and Knowledge Exams}

\section{Introduction}

Large Language Models (LLMs), such as GPT-4 and Llama 2, have made
significant strides in showcasing remarkable capabilities across a wide range of
domains and tasks \citep{touvron2023llama}. These models exhibit a level of
intelligence that surpasses prior Artificial Intelligence (AI) models, excelling in various fields
like coding, medicine, law, agriculture, and psychology, often without
requiring special prompts. Their performance is impressively close to human
expertise, positioning them as potential early versions of artificial general
intelligence (AGI) systems \citep{bubeck2023sparks}. The technology behind LLMs,
which typically involves advanced deep learning techniques, large-scale
transformers, and massive amounts of data, has fueled their rapid progress.
As research on LLMs continues, it is vital to uncover their limitations and
address the challenges in achieving more comprehensive AGI systems. Furthermore,
the machine learning community must evolve beyond traditional benchmarking
methods, moving towards evaluating LLMs in a manner that closely resembles the
assessment of human cognitive abilities. In recent years, significant
advancements in LLMs based on the transformer architecture
\citep{vaswani2017attention} and trained in a self-supervised manner have led to
substantial improvements \citep{nori2023capabilities, devlin2019bert, radford2018improving} in various natural language processing tasks. These
models, trained on massive cross-disciplinary corpora, have the potential to
offer valuable insights and assistance in specialized domains like agriculture,
crop management, and advising. %\citep{liakos2018machine}. 

Most prior work on AI in Agriculture has focused on extracting intelligence on satellite imagery and sensor data. In our research at Microsoft, through Project FarmBeats~\citep{farmbeats, farmbeats1} and Project FarmVibes~\citep{farmvibes}, we have focused on artificial intelligence techniques to drive the adoption of data-driven agriculture. We use AI to clean data sources, for example,  by seeing through clouds~\citep{SpaceEye}, to better data collection by predicting sampling locations, to making better predictions, for example of yields, microclimates~\citep{deepmc}, and carbon sequestration~\citep{farmvibes-causal}. While these advances have helped improve the state of the art in agricultural AI, tech-savviness of farmers and farm workers has limited the adoption of technology solutions. 

In this paper, we look at the ability of LLMs to bridge the technology adoption gap in agriculture. Our first consideration is that of an agronomist's assistant (a virtual agronomist). The vision is that an agronomist can ask questions to the virtual agronomist to remove ambiguity in the response. This can help better customize the answers for the farmer. Such a virtual agronomist can help an agronomist stay up to date with the latest information, can help remove paperwork, and can help them be more effective in their consultation with the farmer. 

We build a virtual agronomist leveraging the advances in LLMs. Our goal is to leverage the LLM to pass an exam that an agronomist needs to pass, for example, the Certified Crop Advisor related Exams in the USA.

%Measuring the ability of LLMs to assist agronomists in
%solving complex agronomy-related problems is crucial for understanding their
%potential in revolutionizing the agricultural sector. As these advanced models
%continue to improve and demonstrate human-like capabilities, it is essential to
%assess their performance in real-world scenarios and their capacity to provide
%practical, innovative solutions to pressing agricultural challenges. To date,
%there has been no comprehensive evaluation of how well LLMs can address
%important problems related to agriculture. Conducting such assessments is key to
%unlocking their full potential and ensuring that agronomists and other
%professionals in the field can leverage the power of artificial intelligence to
%drive sustainable and efficient agricultural practices, ultimately contributing
%to global food security and environmental preservation. Traditional
%evaluation methods may no longer suffice for these advanced models,
%necessitating the use of human evaluation approaches, such as exams and
%certification tests, to assess their capabilities more accurately.  By
%subjecting LLMs to these rigorous assessments, we can gain valuable insights
%into their strengths and limitations and better gauge their potential
%contributions to specific-domain tasks.

We first utilized diverse datasets to estimate the performance of
Large Language Models (LLMs) in the field of agriculture. First, the Brazilian
dataset from Embrapa's database comprises a variety of complex, open
agriculture-related questions and challenges \citep{embrapa500}. Second, we used
questions from the American Society of Agronomy related to the Certified Crop
Adviser (CCA), designed to evaluate an individual's ability to provide competent
advice on crop management, soil, and water conservation \citep{cca2022}. Lastly,
the Indian dataset from AgriExam offers a collection of previous year questions
from various agriculture-related exams for graduate programs, covering topics such as agronomy,
horticulture, and soil science \citep{agriexam2022}. These resources, featuring
multiple-choice questions and short-answer questions
enable us to assess the capabilities of LLMs in answering difficult
agriculture-related questions and explore their potential applications in
providing expert advice and support within the agricultural sector.  

By using
out-of-the-box models, this study assesses how these LLMs can be applied to
provide valuable insights and assistance in the specialized domain of
agriculture.  These findings contribute to the understanding of LLMs' potential
in the agriculture domain and pave the way for their responsible and efficient
use in addressing agricultural challenges and promoting sustainable practices.

In this paper, we make several noteworthy contributions to the understanding of
LLMs in the agriculture domain.  These contributions can be itemized as follows:

\begin{itemize}
\item We present a comprehensive evaluation of LLMs, such as Llama2-13B, Llama2-70B, GPT-3.5 and GPT-4,
in answering agriculture-related questions using benchmark datasets from major
agriculture producer countries.
\item We investigate the impact of retrieval techniques and prompting ensemble methods on the performance of these
models, providing valuable insights into their capabilities and limitations.
\item We discuss the implications of these findings in terms of potential uses
of LLMs in agricultural education, assessment, and practice,
emphasizing the importance of addressing challenges related to accuracy and
safety while harnessing the power of LLMs effectively and responsibly in the
agriculture domain.
\end{itemize}
\section{Related Works}
 
In this section, we review the relevant literature and studies that have
explored the application of artificial intelligence, particularly large language
models, in agriculture and related domains, as well as in educational settings
and assessment methodologies. We present a brief introduction to each paper,
discussing the contributions and potential improvements in each work, and
comparing them to the proposed paper.

\cite{nori2023capabilities} evaluated GPT-4 on medical competency examinations and
benchmark datasets, such as the USMLE and MultiMedQA suite. Despite not being
specialized for medical problems, GPT-4 demonstrated impressive capabilities in
natural language understanding and generation. The study explored various
aspects of model performance, including text and images in test questions,
memorization during training, and probability calibration. Results showed GPT-4
outperforming earlier models and models fine-tuned on medical knowledge, with
improved calibration. The authors discuss GPT-4's ability to explain medical
reasoning, personalize explanations, and craft counterfactual scenarios,
considering implications for medical education and practice, while acknowledging
accuracy and safety challenges.  

\cite{nunes2023evaluating} explored the capabilities of language models (LMs),
specifically GPT-3.5 and GPT-4, in tackling high-stakes multiple-choice tests
using the National High School Exam (Exame Nacional do Ensino Médio - ENEM), a
multidisciplinary entrance examination widely adopted by Brazilian universities.
The ENEM challenges LLMs due to its questions spanning multiple fields of
knowledge and requiring understanding of information from diverse domains. The
authors analyzed responses generated by GPT-3.5 and GPT-4 models for questions
presented in the 2009-2017 exams, as well as for questions of the 2022 exam,
which were made public after the training of the models was completed. Different
prompt strategies were tested, including the use of Chain-of-Thought (CoT)
prompts to generate explanations for answers. The best-performing model, GPT-4
with CoT, achieved an accuracy of 87\% on the 2022 edition, surpassing GPT-3.5 by
11 points.  

In the context of applying Large Language Models (LLMs) for question-answering tasks, a notable study is the Med-PaLM 2 paper by \cite{singhal2023expertlevel}. This model leverages base LLM improvements, medical domain fine-tuning, and prompting strategies, including a novel ensemble refinement (ER) approach that is also used in this research. The model has achieved remarkable results, scoring up to 86.5\% on the MedQA dataset, while also demonstrating high performance on clinical topics datasets.

\cite{dikli2006overview} paper, titled "An overview
of automated scoring of essays," discusses the state of research in automated
essay grading (AEG) and its potential applications in educational settings.
While it provides a comprehensive review of AEG methodologies and their
implications for education, it focuses on a specific assessment method.  "Show
your work: Improved reporting of experimental results" by \cite{dodge2019show}
emphasizes the importance of improved reporting of experimental results. The
work contributes to more transparent and reproducible research practices in the
field of natural language processing.  In "Using Latent Semantic Analysis to
evaluate the contributions of students in AutoTutor," \cite{graesser2000using}
investigated the use of Latent Semantic Analysis (LSA) for evaluating students'
contributions in an intelligent tutoring system. The work focuses on AI-driven
educational systems and their evaluation but is limited in scope, as it is
centered around the AutoTutor system and LSA.

Retrieval-Augmented Generation (RAG) techniques have been gaining attention for
their ability to enhance the performance of large language models in various
domains \citep{lewis2020retrieval}. RAG techniques combine information retrieval
and generation capabilities, allowing models to access and utilize relevant
external knowledge to generate more accurate and informed responses.
Specifically, RAG techniques employ a two-step process: first, retrieving
relevant documents or passages from a knowledge source, and second, conditioning
the generation process on the retrieved information. This approach has
demonstrated success in improving the performance of language models in tasks
such as question-answering and summarization, where access to external knowledge
can be crucial for generating correct and coherent outputs.
\cite{kamilaris2018deep} conducted a survey on "Deep learning in agriculture,"
outlining the use of deep learning techniques in various agricultural tasks. The
paper provides a comprehensive review of deep learning applications in
agriculture.  "Crop yield
prediction using deep neural networks" by \cite{khaki2019crop} explores the use
of deep learning techniques to predict crop yields with high accuracy.
\cite{liakos2018machine}  presented a review titled "Machine learning in
agriculture," discussing various machine learning techniques for addressing
agricultural challenges. The paper provides a comprehensive review of machine
learning applications in agriculture but does not focus on large language models
like GPT.  

\cite{zhao2023chatagri} have explored ChatGPT's potential in
agricultural text classification, resulting in a solution called ChatAgri.
Evaluations on multi-linguistic datasets show competitive performance compared
to existing PLM-based fine-tuning approaches. ChatAgri exhibits strengths in
zero-shot learning, domain transferability, and minimal hardware requirements,
suggesting its suitability as a low-cost AI technique for smart agriculture
applications. The study provides a comprehensive analysis of the findings,
contributing to future sustainable smart agriculture development. 
The paper "WINOGRANDE: An Adversarial Winograd Schema Challenge
at Scale" by \cite{sakaguchi2021winogrande} introduces a large-scale dataset for
evaluating the common sense reasoning capabilities of AI models. This work
contributes to the creation of a challenging benchmark for AI systems, fostering
research in natural language understanding and common sense reasoning.  "XLNet:
Generalized Autoregressive Pretraining for Language Understanding," proposed by
\cite{yang2019xlnet}, presents a novel pretraining method for natural language
understanding tasks. This work develops a competitive language model that
outperforms other methods on several benchmarks.  

In summary, while the existing literature has demonstrated the potential of
artificial intelligence, particularly LLMs, in various domains
and applications, the specific context of agriculture has been less explored.
Our proposed paper distinguishes itself from the reviewed works by focusing on
employing large language models, such as GPT-4, within the agriculture domain.
We aim to evaluate how these models can enhance human capabilities in the
context of agriculture, including answering agriculture-related questions and
providing guidelines for crop management. By bridging the gap between the
advancements in large language models and the agricultural sector, our study
offers a unique perspective and contributes to the understanding of AI's
potential impact on agriculture by providing a baseline for future benchmarks
about the use of large language models to solve agricultural problems.
\section{Datasets}
\label{sec:datasets}

This study assesses advanced language models in addressing
agriculture-related questions by utilizing three distinct datasets with
different formats, each originating from three major crop producer countries.

\subsection{Certified Crop Adviser (CCA) Exam}

\begin{listing}[h]  
\begin{mdframed}[linewidth=1pt,linecolor=black,backgroundcolor=gray!20]  
\centering  
\begin{lstlisting}  
During which stage of plant growth is an above-normal rainfall conducive to pod and stem 
blight as well as Phomopsis seed decay?  
  
a) Flowering  
b) Pod filling  
c) Early vegetative  
d) Mid vegetative  
  
Answer: b) Pod filling  
\end{lstlisting}  
\end{mdframed}  
\caption{Adapted CEU question example, requiring specialized knowledge about Pod  
and Stem Blight and Phomopsis Seed Decay of Soybean.}  
\label{fig:ceu_example}  
\end{listing}  

The Certified Crop Adviser (CCA) Exam is a comprehensive evaluation program
designed to assess the competency of agricultural professionals in agronomy and
crop advising \citep{cca2022}. The CCA certification ensures that crop advisers possess the
necessary knowledge and skills to provide accurate and reliable advice to
farmers and other stakeholders in the agricultural sector. The certification
process involves passing two exams: the International CCA Exam, which focuses on
general agronomic principles, and a Local Board Exam tailored to specific
regional conditions and regulations. To maintain their CCA certification, crop
advisers are required to participate in continuing education activities and earn
Continuing Education Units (CEUs) credits. These CEUs demonstrate that the
advisers are staying up to date with the latest developments in agronomy and are
committed to lifelong learning. A specific number of CEUs must be earned within
a designated time frame to maintain the certification. The questions used for
this evaluation are based on Continuing Education Unit (CEU) materials and not
the actual Certified Crop Adviser (CCA) certification tests. The purpose of
using CEU questions in this study is solely to assess the GPT models' ability to
provide valuable insights and assistance in the specialized domain of
agriculture, and not to replicate or undermine the CCA certification process.

The Crop Protection Network (CPN) is an excellent resource for CCAs seeking to
earn CEUs. The CPN offers a variety of online quizzes related to continuing
education units'
credits\footnote{https://cropprotectionnetwork.org/cca-ceu-quizzes}. These
quizzes cover a wide range of topics in agronomy and crop management, such as
disease management, pest control, nutrient management, and environmental
stewardship. By participating in these quizzes and earning CEUs, CCAs can ensure
that they remain knowledgeable and competent in their field, providing the best
possible guidance to farmers and contributing to the overall sustainability and
productivity of the agricultural sector.

The adapted CEU question example provided in Listing \ref{fig:ceu_example}
highlights the specialized knowledge required to answer questions related to
specific crop diseases, such as pod and stem blight and phomopsis seed decay of
soybean. 

\subsection{EMBRAPA Dataset}

\begin{listing}[h]
\begin{mdframed}[linewidth=1pt,linecolor=black,backgroundcolor=gray!20]
\centering
\begin{lstlisting}
What are the symptoms and damages caused by the gall nematode in citrus plants?

Answer: The symptoms of the gall nematode in citrus plants include dysfunctionin the roots,
leading to smaller leaf size and marked chlorosis. This resultsin a reduction in production.
The growth of the main root may be paralyzed,causing a proliferation of lateral roots. In
severe infections, the plant maydie before the nematode completes its cycle. However, not all
citrus species are susceptible to this nematode, so it is essential to identify the species
and monitor it in the soil and roots.
\end{lstlisting}
\end{mdframed}
\caption{Example of an adapted question from the Embrapa dataset, focusing on
the symptoms and damages caused by the gall nematode in citrus plants.}
\label{fig:gall_nematode_example}
\end{listing}

In this study, we use an extensive dataset, "500 Perguntas 500 Respostas -
Embrapa/SCT" (500 Questions 500 Answers - Embrapa/SCT), provided by the
Brazilian Agricultural Research Corporation (Embrapa) \citep{embrapa500}. This
dataset consists of a series of books containing questions and answers related
to various aspects of crop cultivation and management in Brazil. These questions
have been formulated by a diverse group of stakeholders, including producers,
farmers, and farming associations, and are accompanied by responses from Embrapa
specialists. Unlike CCA exams, the questions in this dataset are not
presented in a multiple-choice format but as general text-based inquiries,
requiring a deeper understanding of the subject matter to generate accurate and
relevant answers.

The use of this robust dataset presents a unique opportunity to examine the
effectiveness of advanced language models in answering
domain-specific questions accurately and contextually. By rigorously evaluating
LLM's performance on the Embrapa dataset, our goal is to explore the potential
of artificial intelligence in providing reliable and efficient solutions to
real-world agricultural challenges. Through this analysis, we aim to bridge the
gap between state-of-the-art technology and sustainable agricultural practices,
ultimately contributing to the improvement of agricultural productivity and
resource management on a global scale.

Listing \ref{fig:gall_nematode_example} illustrates an example of an adapted
question from the Embrapa dataset, demonstrating the challenges of verifying the
correctness of answers in this context. Since the dataset's questions are not
multiple-choice and often have no single correct answer, assessing the accuracy
of responses can be subjective. This highlights the importance of understanding
the nuances of agricultural topics and the expertise required to evaluate the
relevance and correctness of AI-generated answers. The Embrapa dataset,
translated as "500 Questions 500 Answers," not only targets agronomists but also
serves as a valuable resource for farmers. It encompasses a wide range of topics
essential to both groups, contributing to the overall knowledge base and
fostering sustainable agricultural practices.

\subsection{AgriExams Questions}

\begin{listing}[h]
    \begin{mdframed}[linewidth=1pt,linecolor=black,backgroundcolor=gray!20]
  \centering
\begin{lstlisting}
Which directorate/institute of ICAR coordinates/organizes Parthenium Awareness
Week in India during the month of August each year?

a) ICAR-DWR, Jabalpur
b) ICAR-IARI, New Delhi
c) ICAR-CSSRI, Karnal
d) ICAR-IIFSR Modipuram

Answer: a) ICAR-DWR, Jabalpur
\end{lstlisting}
\end{mdframed}
  \caption{Adapted AgriExam question example, highlighting the multiple-choice
  style and focus on Indian agriculture.}
  \label{fig:agriexam_example}
\end{listing}

The AgriExam dataset features a wide range of questions from various
agriculture-related exams, primarily focusing on Indian agriculture
\citep{agriexam2022}. These questions are collected from different entrance
exams, including those for prestigious institutions like JNKVV (Jawaharlal Nehru
Krishi Vishwa Vidyalaya) and RVSKVV (Rajmata Vijayaraje Scindia Krishi Vishwa
Vidyalaya). The questions are designed to assess candidates' knowledge and
understanding of diverse agricultural topics, ranging from crop production and
soil science to agricultural economics and extension, with a specific focus on
India's farming practices and conditions.

Listing \ref{fig:agriexam_example} shows an example of a multiple-choice
question style adapted from the AgriExam dataset. Note that this is not the
exact question from the source but serves as a representation of the question
style.  The questions 
often structured as multiple-choice or descriptive questions, requiring
test-takers to demonstrate a deep understanding of the subject matter. In this paper, we 
evaluated only the Agriexams multiple-choice questions.
Additionally, the questions may focus on the application of scientific
principles to real-world agricultural challenges, particularly those relevant to
Indian agriculture, thereby encouraging problem-solving and critical thinking
skills.
\section{Method Overview}

\begin{figure}[htbp]
  \centering
  \includegraphics[width=0.7\textwidth]{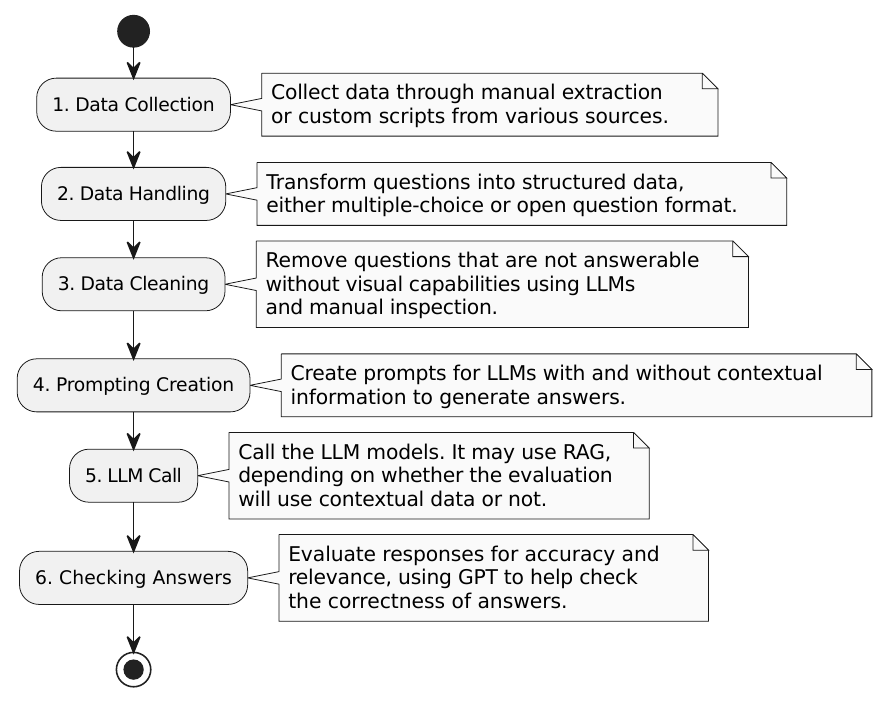}
  \caption{Answering and Evaluation Scheme}
  \label{fig:answering_method}
\end{figure}

\begin{comment}
@startuml
actor "User" as user

entity "Orchestrator" as orchestrator

entity "Answer Agent" as answer_agent  

entity "Evaluation Agent" as eval_ans  

entity "LLM" as llm

user --> orchestrator : Raw Questions  

orchestrator --> answer_agent : Raw Question 

answer_agent --> llm : Question Prompt

llm --> orchestrator: LLM Answer

orchestrator --> eval_ans : Question \n+ LLM Answer \n+ Ground Truth Answer

eval_ans --> llm : Evaluation Prompt

llm --> orchestrator: Score (0 or 1)

orchestrator --> user: Score (0 or 1)
@enduml
\end{comment}

This section outlines the method that was employed to execute experiments using LLMs to answer agriculture-related questions (Figure \ref{fig:answering_method}).

\textbf{1. Data Collection:} The first step in the process involved collecting the data required for the experiments. This was done through manual extraction of questions and answers from various sources, such as books, articles, or online resources. Alternatively, custom scripts were developed to automate the data collection process, particularly when dealing with large datasets or multiple sources.

\textbf{2. Data Handling:} Once the data was collected, it needed to be transformed from text into structured data. This was done by converting the questions into either a multiple-choice format using JSON or an open question format, depending on the nature of the questions and the desired output of the experiment. Structuring the data in this way allowed for easier integration with the LLMs and facilitated subsequent analysis of the results.

\textbf{3. Data Cleaning:} The next step was data cleaning, which involved removing questions that were not answerable without visual capabilities, such as questions based on images. To clean the data, both LLMs and manual inspection were used. This step ensured that the remaining questions were suitable for the LLMs and would yield meaningful results in the context of agriculture-related inquiries.

\textbf{4. Prompting Creation:} With the cleaned and structured data in hand, the next step was to create prompts for the LLMs to generate answers to the questions. The prompts were designed to include contextual information, such as background knowledge about the subject matter, to improve the accuracy and relevance of the generated responses. This step was crucial in guiding the LLMs to produce high-quality answers that addressed the specific needs of the agriculture-related questions.

\textbf{5. LLM Call:} In this step, the LLMs were called and could use RAG and/or ER depending on the contextual data availability and computational resources budget. These models generated responses to the prompts created in the previous step.

\textbf{6. Checking Answers:} Finally, after the LLMs generated answers to the questions, it was essential to check and evaluate the responses for accuracy and relevance. This step involved comparing the generated answers to the original answers provided by experts, as well as assessing the overall quality of the responses in terms of their applicability to the agriculture domain. In this step, LLMs were utilized to help check the correctness of answers, especially for open questions.

\subsection{Prompting Strategy}

\begin{figure}[htbp]
  \centering
  \includegraphics[width=0.7\textwidth]{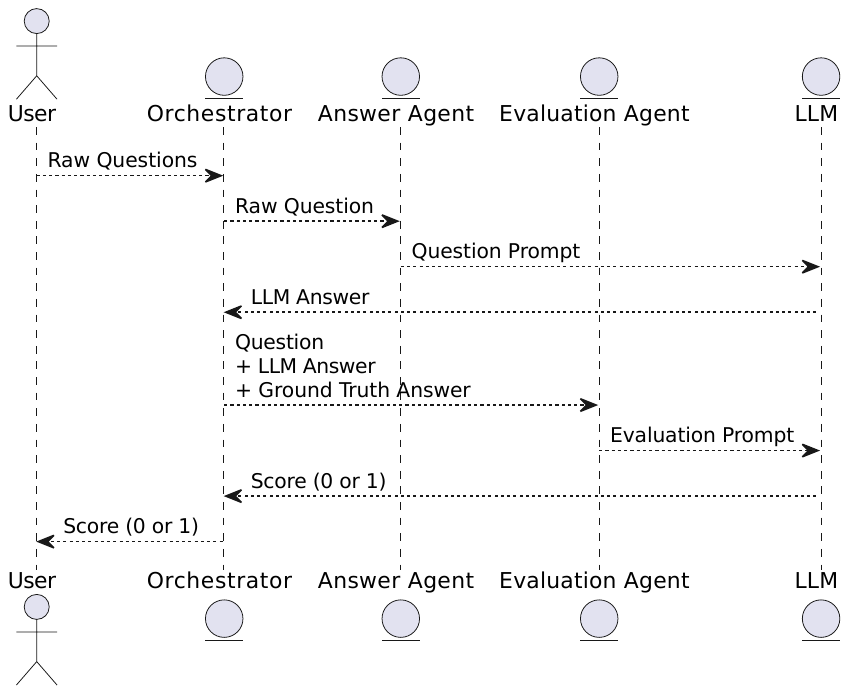}
  \caption{Prompting Scheme}
  \label{fig:prompting_method}
\end{figure}

\begin{comment}

@startuml
!define AWSPUML https://raw.githubusercontent.com/awslabs/aws-icons-for-plantuml/v14.0/Advanced/PlantUML    
  
skinparam monochrome true    
  
start    
  
:1. Data Collection;    
note right    
  Collect data through manual extraction    
  or custom scripts from various sources.    
end note    
  
:2. Data Handling;    
note right    
  Transform questions into structured data,    
  either multiple-choice or open question format.    
end note    
  
:3. Data Cleaning;    
note right    
  Remove questions that are not answerable    
  without visual capabilities using LLMs    
  and manual inspection.    
end note    
  
:4. Prompting Creation;    
note right    
  Create prompts for GPT with and without contextual    
  information to generate answers.    
end note    
  
:5. LLM Call;  
note right  
  Call the LLM models. It may use RAG,  
  depending on whether the evaluation  
  will use contextual data or not.  
end note  
  
:6. Checking Answers;    
note right    
  Evaluate responses for accuracy and    
  relevance, using GPT to help check    
  the correctness of answers.    
end note    
  
stop
@enduml
\end{comment}

\begin{listing}[h]
    \begin{mdframed}[linewidth=1pt,linecolor=black,backgroundcolor=gray!20]
  \centering
\begin{lstlisting}[language=Java] 
template_question = """ 

{preamble}

Answer the following question and provide the correct answer. 

The question is a multiple-choice question with a unique correct answer. 

{question} 

Answer the question by providing the correct alternative (example: the correct
option is b). 

Do not provide an empty answer. 

""" 
\end{lstlisting}
\end{mdframed}
  \caption{GPT question answering prompt.}
  \label{fig:question_prompt}
\end{listing}

\begin{listing}[h]
    \begin{mdframed}[linewidth=1pt,linecolor=black,backgroundcolor=gray!20]
  \centering
\begin{lstlisting}[language=Java]
template_evaluation = """ 

You are an expert professor specialized in checking students' answers to
questions. 

You are checking the following question: 

{query} 

Here is the real answer: 

{answer} 

You are checking the following student's answer: 

{result} 

What grade do you give, where 0 is incorrect and 1 is correct? Give me only 0 or
1 as response. 

If the student's answer is not related to the question, give me 0. 
""" 
\end{lstlisting}
\end{mdframed}
  \caption{GPT answer evaluation prompt.}
  \label{fig:listing_evaluation}
\end{listing}

In this research, we employ two distinct LLM agents to perform separate tasks.
The first agent (Answer Agent) is responsible for answering the questions
provided. The second agent (Evaluation Agent) is assigned the task of evaluating
the correctness of the answers generated by the first agent. This dual-agent
setup allows for a more thorough examination of the AI models' performance, as
one focuses on generating responses while the other ensures the accuracy and
relevance of the answers produced. We also use the Orchestrator to manage the
message passing between the models.

Figure \ref{fig:prompting_method}, shows the prompting process. Initially, the User submits raw questions
to the Orchestrator, which then presents a raw question to the Answer Agent. The
Answer Agent creates a question prompt for the language model, and the LLM
processes the prompt, generating an answer sent back to the Orchestrator. Listing \ref{fig:question_prompt}
shows the question prompt used by the Answer Agent and passed to the LLM. In the listing, the placeholder \texttt{\{preamble\}} 
is used to pass some contextual information (e.g., exam name and/or location) to the answering agent while
the placeholder
\texttt{\{question\}} is replaced with the user questions. 

Next, the Orchestrator forwards the original question, LLM's answer, and the ground
truth answer to the Evaluation Agent. Subsequently, the Evaluation
Agent formulates an evaluation prompt and delivers it to the LLM.  The
evaluation prompt used by the Evaluation Answer is presented in Listing
\ref{fig:listing_evaluation} where \texttt{\{query\}} is placeholder for the
question, \texttt{\{answer\}} is the ground truth question, and
\texttt{\{result\}} is the answer provided by the Answer Agent.  The LLM
evaluates the prompt and returns a score (either 0 or 1) to the Orchestrator.
Finally, the Orchestrator communicates the score to the User, concluding the
process.

\subsection{Context Provisioning}

In this study, we used the Llama Index \citep{liu2022llama_index} to create
context-based questions for LLMs when addressing questions related to
agricultural applications, specifically for USA and India datasets. 
As Embrapa's dataset is composed of open questions, the process of providing 
context is not applicable as the answer for the questions would be directly
presented in the background material. Remember Embrapa's dataset is composed
of direct questions and answers related to Brazilian agriculture.

The Llama Index organized and indexed the same text material provided to certified agronomists
when obtaining credits to renew their certificates. By incorporating the Llama
Index, we aimed to improve the LLMs' performance on domain-specific tasks
by aligning the context of a given question with the knowledge embedded in the
models.  The Llama Index organizes and indexes a large collection of
domain-specific information, like agricultural texts, research articles, and
expert opinions, into a structured and easily accessible format. This indexed
information can then be used to create context-based questions for the LLMs,
providing them with the necessary background knowledge to generate informed and
accurate responses to questions in the agriculture domain.  To generate a
context-based question using the Llama Index, we extract important contextual
information from the identified resources, select key facts, and
concepts directly related to the questions.  
Finally, we crafted context-based questions by combining the original questions
with the extracted contextual information, ensuring that the GPT models had
access to the necessary background knowledge for generating accurate responses. 

\subsection{Ensemble Refinement Method}

Ensemble Refinement (ER) is a prompting strategy proposed by \cite{singhal2023expertlevel} that builds on techniques like chain-of-thought and self-consistency \citep{wei2023chainofthought, wang2023selfconsistency}, leveraging the idea of conditioning a LLM on its own generations before producing a final answer. ER is a two-stage process designed to improve the performance of the LLM when answering questions, particularly in the context of agriculture-related questions.

In the first stage of ER, a (few-shot) chain-of-thought prompt and a question are given to the model. The model then generates multiple possible explanations and answers stochastically via temperature sampling. Each generation consists of an explanation and an answer for a question. In the second stage, the model is conditioned on the original prompt, the question, and the concatenated generations from the previous step. The model is then prompted to produce a refined explanation and answer. This can be viewed as a generalization of self-consistency, where the LLM aggregates answers from the first stage instead of conducting a simple vote. By considering the strengths and weaknesses of the generated explanations, the LLM can produce more refined answers.

To further improve performance, the second stage is performed multiple times. The final answer is then determined by a plurality vote over the generated answers.
In this paper, due to the resource cost associated with repeated samplings from a model, ER is only applied for multiple-choice evaluation.
\section{Results}

This section presents the results for establishing a baseline for LLMs in agriculture, evaluating their 
performance on various datasets and question types. Our findings provide 
insights into LLM strengths and limitations, guiding future research and 
novel strategies for enhancing their capabilities. Table~\ref{table:dataset_sizes} presents the number
of questions and the adopted preambles for each dataset.
In our experiments, we used Azure Open AI deployments for GPT-3.5 and GPT-4. Additionally, for inference, we employed Llama2 models, specifically 13B on a single Nvidia-H100 and 70B on two Nvidia-H100s.

\begin{table}[h]
\centering
\begin{tabularx}{\textwidth}{lcp{9.6cm}}
\hline
Dataset & \multicolumn{1}{c}{\#Questions} & \multicolumn{1}{c}{Preamble} \\
\hline
CCA Video-Based Questions & 89 & The following question is related to agriculture and is part of the process of earning credits that are necessary for the renewal of the Certified Crop Adviser (CCA) certification. It is question related to the agricultural industry in the United States. \\
CCA Text-Based Questions & 312 & Same as Previous preamble. \\
Embrapa & 18,955 & The following question is related to Brazilian agriculture and is part of a series of questions and answers for Brazilian farmers distributed by Embrapa.\\
AgriExams & 1,723 & The following question is related to Indian agriculture and is part of an exam to ingress in an important graduate program from an Indian university. \\
\hline
\end{tabularx}
\caption{Number of questions and preambles for each dataset.}
\label{table:dataset_sizes}
\end{table}

\subsection{USA Dataset Results}

In our evaluation, we split the CCA-related questions into two categories: video-based
questions and text-based questions. For the text-based questions, we assessed
the performance of LLMs both with and without the use of
the RAG, ER, and using or not a preamble.  Just like in the original ER paper, we adopted for all experiments 11 samplings for the first stage and 33 samplings for the second stage.
For video-based questions, we evaluated LLM performance without
using retrieval techniques, as there is no textual information available in the
videos to serve as context. 
This approach helps us understand how well LLMs can
answer agriculture-related questions based solely on their pre-trained
knowledge, without relying on external context provided by RAG techniques.

\begin{table}[h]  
\centering  
\begin{tabular}{lccccccc}      
\hline  
Scenario & \multicolumn{1}{c}{ER} & \multicolumn{1}{c}{Preamble} &  \multicolumn{1}{c}{Llama2-13B} & \multicolumn{1}{c}{Llama2-70B} &   \multicolumn{1}{c}{GPT-3.5} & \multicolumn{1}{c}{GPT-4} \\  
\hline  
1        &                 &             & 55\%                         & 74\%                      & 74\%                         & 79\%                      \\  
2        &                 &$\checkmark$ & 58\%                         & 72\%                      & 74\%                         & 83\%                      \\  
3        &  $\checkmark$   &$\checkmark$ & --                         & --                      & 80\%                         & 83\%                      \\  
       
\hline  
\end{tabular}  
\caption{Video-based questions results.}
\label{table:video_results_cca}
\end{table}  

The results presented in Table \ref{table:video_results_cca} demonstrate the performance of Llama2-13B, Llama2-70B, GPT-3.5, and GPT-4 on video-based questions under various conditions. In the baseline scenario (Scenario 1), GPT-4 exhibited the best performance, achieving a score of 79\%, followed by GPT-3.5 and Llama2-70B at 74\%, and Llama2-13B at 55\%. When a preamble was introduced in Scenario 2, GPT-3.5's performance remained unchanged at 74\%, while GPT-4's score increased to 83\%, indicating the effectiveness of a preamble in boosting GPT-4's performance. In this scenario, Llama2-13B and Llama2-70B scored 58\% and 72\%, respectively. In our trials, GPT-3.5 and GPT-4 consistently exceeded the performance of Llama2-13B and Llama2-70B. Consequently, to reduce evaluation costs and focus our efforts on the most promising outcomes, we opted to employ ensemble refinement techniques solely on GPT-3.5 and GPT-4, for all experiments in this paper. In Scenario 3, ER was applied to GPT-3.5 and GPT-4. For GPT-3.5 and GPT-4, both models experienced improved scores, with GPT-3.5 achieving 80\% and GPT-4 maintaining its 83\% score. The findings suggest that GPT-4 consistently outperforms other models across all scenarios, with further enhancements observed when incorporating a preamble or ER.

\begin{table}[h]
\centering
\begin{tabular}{lcccccccc}
\hline
Scenario & \multicolumn{1}{c}{RAG}  & \multicolumn{1}{c}{ER} & \multicolumn{1}{c}{Preamble} &  \multicolumn{1}{c}{Llama2-13B} & \multicolumn{1}{c}{Llama2-70B} &   \multicolumn{1}{c}{GPT-3.5} & \multicolumn{1}{c}{GPT-4} \\  
\hline
1 & & & &  47\%   & 55\%   & 64\% & 79\% \\
2 & & & $\checkmark$ & 48\%     & 60\%    & 66\% & 82\% \\
3 & & $\checkmark$ & $\checkmark$ &  --   & --   & 71\% & 84\% \\
4 & $\checkmark$ & & & 70\%    & 81\%   & 82\% & 93\% \\
5 & $\checkmark$&  & $\checkmark$ & 71\%    & 81\%    & 88\% & 93\% \\
\hline
\end{tabular}
\caption{Multiple choice questions results (CCA-related exams - text based).}
\label{table:text_results_cca_five_scenarios}
\end{table}

The results from Table \ref{table:text_results_cca_five_scenarios} demonstrate the performance of Llama2-13B, Llama2-70B, GPT-3.5, and GPT-4 in answering text-based multiple-choice questions from CCA-related exams under various scenarios. GPT-4 consistently outperforms all other models in every scenario, with a noticeable increase in performance when provided with assistance. For instance, when a preamble was introduced in Scenario 2, GPT-4's performance increased from 79\% to 82\%. The combination of ER and a preamble in Scenario 3 further enhanced GPT-4's score to 84\%.

Interestingly, the most significant improvement for all models was observed when employing the RAG (Retrieval-Augmented Generation) technique in Scenarios 4 and 5. GPT-4 achieved an impressive 93\% score, while Llama2-13B and Llama2-70B both saw substantial improvements, reaching 70\% and 81\%, respectively. GPT-3.5's performance also improved, reaching 82\% in Scenario 4 and 88\% in Scenario 5.

While the contextual information provided in preambles is static, RAG retrieves information dynamically, allowing for better adaptation to the context of the questions and potentially resulting in more accurate responses. ER also improves results by harnessing hidden knowledge within LLMs using different temperatures for each GPT call, enabling the extraction of diverse and domain-specific insights for more accurate and contextually relevant answers.

For some questions, LLMs give wrong answers 
within the context of agriculture. For instance, 
Listing \ref{fig:cca_error_example} shows an example of how GPT-4 incorrectly answers the question.

\begin{listing}[h]
\begin{mdframed}[linewidth=1pt,linecolor=black,backgroundcolor=gray!20]
\centering
\begin{lstlisting}
Which biotic stress factors have received the least attention in terms of 
commercially released transgenes?  
a) Insects    
b) Diseases    
c) Weeds    
d) Mammals

Answer: b
GPT Answer: d - Mammals
\end{lstlisting}
\end{mdframed}
\caption{Adapted example of incorrect answer from CCA dataset (GPT-4 with preamble and no RAG).}
\label{fig:cca_error_example}
\end{listing}

According to previous results, we can see that both RAG and ER methods can improve the quality of answers generated by LLMs. However, combining these methods to answer each question in the dataset can increase the computational costs for each question-answering task. Therefore, to evaluate the impact of ER and RAG simultaneously on GPT-X, we sampled 100 multiple-choice questions and assessed the models' performance. We also used preambles for the sampled dataset in all scenarios.

\begin{table}[h]  
\centering  
\begin{tabular}{lccccc}      
\hline  
Scenario & \multicolumn{1}{c}{RAG} & \multicolumn{1}{c}{ER} &  \multicolumn{1}{c}{GPT-3.5} & \multicolumn{1}{c}{GPT-4} \\  
\hline  
1        &                 &             & 70\%                         & 85\%                      \\  
2        &                 &$\checkmark$ & 70\%                         & 92\%                      \\  
3        &  $\checkmark$   &             & 92\%                         & 93\%                      \\  
4        &  $\checkmark$   & $\checkmark$  & 92\%                       & 97\%                      \\  
       
\hline  
\end{tabular}  
\caption{RAG + ER Evaluation table (CCA-related exams - sampled text-based questions).}
\label{table:rag_er_text}
\end{table}

Table \ref{table:rag_er_text} shows the results related to incorporating Ensemble Refinement (ER) and Retrieval-Augmented Generation (RAG) methods to the text-based CCA sampled dataset. Our findings indicate that the combination of these techniques significantly improves the quality of answers generated by Large Language Models (LLMs). In particular, GPT-4 demonstrated the highest performance, reaching 97\% accuracy when both RAG and ER were applied. Interestingly, while ER alone showed a substantial impact on GPT-4's performance, increasing accuracy from 85\% to 92\%, it did not have a noticeable effect on GPT-3.5's performance. In contrast, RAG alone greatly improved GPT-3.5's accuracy from 70\% to 92\%, while only slightly increasing GPT-4's accuracy from 85\% to 93\%. These results suggest that incorporating retrieval and ensemble refinement techniques can enhance LLMs' performance, with the combination of RAG and ER yielding the most significant improvements.

While these techniques can significantly improve the quality of answers generated by Large Language Models (LLMs), it is important to note that the combination of RAG and ER comes with increased computational costs for each question-answering task. ER, in particular, involves multiple samplings in its two-stage process, which can be resource-intensive and time-consuming. RAG, on the other hand, combines retrieval with generative models, which may require additional processing time and resources for accessing and integrating external knowledge.
Despite the increased costs, the substantial improvements in model performance, as demonstrated by the results, may justify the use of these techniques in certain applications. However, it is essential to carefully weigh the trade-offs between performance gains and resource requirements when deciding whether to employ RAG, ER, or both in a given context.

\subsection{Embrapa Results}

\begin{table}[h]
\centering
\begin{tabular}{lcccc}
\hline
Scenario & \multicolumn{1}{c}{Preamble} & \multicolumn{1}{c}{GPT-3.5} & \multicolumn{1}{c}{GPT-4} \\
\hline
1 & & 59\% & 72\% \\
2 & $\checkmark$ & 78\% & 84\% \\
\hline
\end{tabular}
\caption{Embrapa questions results.}
\label{table:embrapa}
\end{table}

\begin{listing}[h]
\begin{mdframed}[linewidth=1pt,linecolor=black,backgroundcolor=gray!20]
\centering
\begin{lstlisting}
Question: Some countries have used in the culture of sorghum protectors known as safeners. Can
this technology be used to use herbicides in sorghum crops?

Answer: Although the effect of seed protectors used to protect plants against the phytotoxic
effect of herbicides is known, their use in sorghum is not allowed in Brazil, as it is not
registered on the Map.

GPT Answer: Yes, the use of safeners in sorghum crops can help protect the plants from 
potential damage caused by herbicides. Safeners are chemicals that selectively protect the
crop from the harmful effects of herbicides, allowing for better weed control without harming 
the sorghum plants. This technology has been successfully used in some countries and can be a 
valuable tool for managing weeds in sorghum crops.
\end{lstlisting}
\end{mdframed}
\caption{Adapted example of incorrect answer from Embrapa dataset (GPT-4 with preamble).}
\label{fig:embrapa_error_example}
\end{listing}

Table \ref{table:embrapa} highlights the performance
of GPT-3.5 and GPT-4 on open-ended questions related to Brazilian agriculture.
One challenge in assessing the correctness of the answers generated by the
models for Embrapa questions lies in their open-ended nature, which allows for
multiple valid responses. Moreover, the RAG technique was not applied in these
experiments to ensure a fair comparison, given that the Embrapa dataset already
contains question-answer pairs. General
models like GPT-3.5 and GPT-4 may struggle to provide accurate responses
tailored to the Brazilian context due to potential limitations in their
training data concerning Brazil's unique agricultural conditions and 
practices (e.g., Listing \ref{fig:embrapa_error_example}).
Despite these challenges, the results suggest that GPT-4 is more capable of
understanding and answering open-ended questions related to Brazilian
agriculture than GPT-3.5, even without using RAG and accounting for the
dataset's specific characteristics. Additionally, when a preambled is added to the
questions specifying the fact they are related to the Brazilian agriculture,
we can see an improvement on the accuracy for both models.

\subsection{AgriExams Results}

\begin{table}[h!]
\centering
\begin{tabular}{lccccccc}
\hline
Scenario & \multicolumn{1}{c}{RAG}  & \multicolumn{1}{c}{ER} & \multicolumn{1}{c}{Preamble} &  \multicolumn{1}{c}{Llama2-13B} & \multicolumn{1}{c}{Llama2-70B} &   \multicolumn{1}{c}{GPT-3.5} & \multicolumn{1}{c}{GPT-4} \\  
\hline
1 & & &  &  47\%   & 50\% & 54\% & 65\% \\
2 & & & $\checkmark$  &  45\%   & 48\% & 55\% & 66\% \\
3 & & $\checkmark$ & $\checkmark$  &  --   & -- & 59\% & 70\% \\
4 & $\checkmark$ & &  &  48\%   & 52\% & 68\% & 78\% \\
5 & $\checkmark$ & & $\checkmark$  &  48\%   & 52\% & 68\% & 79\% \\
\hline

\end{tabular}
\caption{AgriExams questions results.}
\label{table:agriexam}
\end{table}

The results from Table \ref{table:agriexam} demonstrate the performance of Llama2-13B, Llama2-70B, GPT-3.5, and GPT-4 in answering AgriExams questions under various scenarios. As observed in the previous experiments, GPT-4 consistently outperforms all other models in every scenario, with noticeable increases in performance when provided with different methods. For instance, when a preamble was introduced in Scenario 2, GPT-4's performance increased from 65\% to 66\%, while Llama2-13B's and Llama2-70B's scores slightly decreased. Sometimes, Llama2-13B and Llama2-70B may not be as effective at utilizing the contextual information provided by the preamble as GPT-3.5 and GPT-4. The latter models have been trained on larger datasets and possess more advanced architectures, which may enable them to better understand and integrate contextual information in their responses.

The combination of Expert Refinement (ER) and a preamble in Scenario 3 further enhanced GPT-4's score to 70\%. Interestingly, the most significant improvement for GPT-3.5 and GPT-4 was observed when employing the RAG (Retrieval-Augmented Generation) technique in Scenarios 4 and 5, with GPT-4 achieving scores of 78\% and 79\%, respectively. Llama2-13B and Llama2-70B also experienced improvements in Scenarios 4 and 5, reaching scores of 48\% and 52\%, respectively. These findings underscore the importance of incorporating additional information and techniques, such as a preamble, ER, and RAG, to enhance the performance of large-scale language models in specialized question-answering tasks, such as those found in AgriExams.

Considering that the AgriExams questions are tailored to India's agriculture,
the models may face challenges when answering questions specific to the region.
Additionally, the background material
provided\footnote{\url{https://www.agriexam.com/agriculture-pdf-notes}} is not
explicitly designed for these questions, unlike the USA dataset. They are general
guidelines related to Indian's agriculture which may help to answer the questions
or not depending on how they are formulated.
Listing~\ref{fig:agriexam_error_example} shows an example of question 
that was not able to be responded correctly by GPT-4 model.

\begin{listing}[h]
\begin{mdframed}[linewidth=1pt,linecolor=black,backgroundcolor=gray!20]
\centering
\begin{lstlisting}
Who is the advocate of the "Marthandam" Rural Development Programme?
a) Spencer Hatch
b) Daniel Hamiltion
c) A.T. Mosher
d) M.K. Gandhi

Answer: a
GPT Answer: b) M.K. Gandhi
\end{lstlisting}
\end{mdframed}
\caption{Adapted example of incorrect answer from Agriexams dataset (GPT-4 without RAG and with preamble).}
\label{fig:agriexam_error_example}
\end{listing}

\begin{figure}[htbp]
  \centering
  \includegraphics[width=0.7\textwidth]{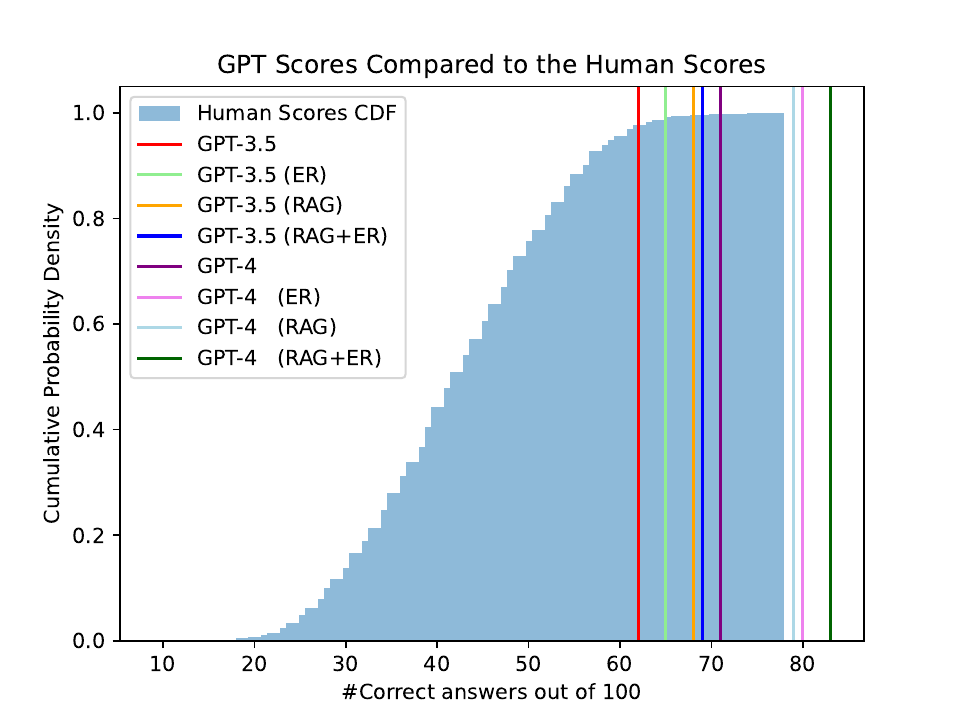}
  \caption{GPT-X results compared to human scores for JNKVV \& RVSKVV M.Sc Entrance exam (2022). Rank represents the hypothetical position of GPT-X on the exam. Percentile
  is the CDF percentile for the hypothetical GPT-X score.}
  \label{fig:human_comparison}
\end{figure}

Just like the CCA-based text exams, we sampled 100 questions from the Agriexams dataset 
to evaluate the GPT-X performance using ER and RAG (Table \ref{table:rag_er_agri}). In this case, we used the questions from the JNKVV \& RVSKVV M.Sc Entrance exam (2022) \footnote{\url{https://www.agriexam.com/jnkvv-rvskvv-m-sc-entrance-paper}} which contains multiple choice questions related to Indian agriculture and focused only on GPT-X architectures.
Besides the questions and answers we also had access to the human results for this dataset \citep{kvvmponline}.
Figure \ref{fig:human_comparison} presents the human scores' cumulative distribution
function (CDF)  and the GPT results for versions 3.5 and 4, with and without RAG and ER. The results are also presented in Table \ref{table:rag_er_agri} with scores, ranks and percentile for each scenario of GPT-X evaluation. For each scenario, we just compared GPT-X results to the human available scores.

\begin{table}[h]    
\centering  
\begin{tabular}{lccccccccc}  
\hline    
Scenario & \multicolumn{1}{c}{RAG} & \multicolumn{1}{c}{ER} &  \multicolumn{3}{c}{GPT-3.5} & \multicolumn{3}{c}{GPT-4} \\    
\cline{4-6} \cline{7-9}  
& & & Score & Rank & Percentile & Score & Rank & Percentile \\    
\hline    
1        &                 &             & 62\% & 57 & 97.35\%   & 71\%     & 6 & 99.72\%                  \\    
2        &                 &$\checkmark$ & 65\% & 24 & 98.99\%   & 80\%     & 1 & 100\%                  \\    
3        &  $\checkmark$   &             & 68\% & 11 & 99.55\%   & 79\%     & 1 & 100\%                  \\    
4        &  $\checkmark$   & $\checkmark$  & 69\% & 8 & 99.66\%  & 83\%    & 1 & 100\%                   \\    
         
\hline    
\end{tabular}    
\caption{RAG + ER Evaluation table (JNKVV \& RVSKVV M.Sc Entrance exam (2022).}  
\label{table:rag_er_agri}  
\end{table}

Just like the previous results, we can see GPT-4 consistently outperforms GPT-3.5 across all scenarios, with noticeable improvements when provided with additional methods. In particular, GPT-4 maintains the top rank ER or RAG is introduced. Furthermore, the combination of RAG and ER in Scenario 4 led to the highest performance for both models, with GPT-3.5 reaching a 99.66\% percentile and GPT-4 maintaining a top 1 rank. These findings underscore the value of utilizing multiple techniques, such as RAG and ER, to improve the performance of language models in domain-specific question-answering tasks like the JNKVV \& RVSKVV M.Sc Entrance Exam (2022). In contrast, GPT-3.5 experienced more modest increases in accuracy across all scenarios. When both RAG and ER were applied, its accuracy reached 69\%, only slightly higher than the 68\% achieved when using RAG alone. These results highlight the potential benefits of employing RAG and ER techniques to enhance the performance of Large Language Models (LLMs) in specific contexts, such as the Agriexams sampled dataset. However, the effectiveness of each method may vary depending on the particular model and dataset being used.

\section{Conclusion}

This study aimed to establish a baseline for assessing the capabilities of large
language models, such as LLama 2, GPT-3.5 and GPT-4, in addressing complex
problems in agriculture. By evaluating their performance on various datasets and
question types, the study provides valuable insights into the strengths and
limitations of LLMs within the agricultural domain. It also highlights the
importance of leveraging AI in a responsible and effective manner, addressing
challenges related to accuracy, safety, and potential biases.

The primary contributions of the paper include the establishment of performance
baselines for LLMs on agriculture-related problems, enabling researchers and
practitioners to compare their results with the current state of LLM performance
on these problems. Furthermore, the exploration of the RAG and ER technique's impact on LLMs demonstrates its potential in
improving the performance of LLMs on region-specific questions, although
GPT-4 remains superior even without using RAN and ER.

The results demonstrated that GPT-4 consistently outperforms other models across
different datasets and question types, including video-based, text-based, and
open-ended questions. This superior performance can be attributed to GPT-4's
larger model size and more extensive training dataset, which enables it to
capture complex patterns and relationships more effectively. The findings
emphasize the potential of GPT-4, particularly when combined with retrieval
techniques like RAG, in solving complex agriculture-related problems and
answering questions accurately.

By providing a starting point for further investigation, this study hopes to
foster innovation and collaboration among researchers, ultimately leading to the
effective application of LLMs in agriculture. This, in turn, can contribute to
the development of more efficient and sustainable farming practices, addressing
critical challenges in global food security and environmental conservation.
As we move forward, it is crucial to continue refining our understanding of
GPT-4's capabilities and explore richer prompting strategies, alignment
techniques, and metacognitive capabilities. Additionally, future research should
focus on mitigating the risks associated with erroneous generations and biases
while maximizing the benefits of AI integration in agriculture.
Overall, this study serves as a stepping stone towards unlocking the potential
of AI in agriculture, paving the way for innovative solutions, improved resource
management, and more resilient agricultural systems capable of adapting to the
challenges posed by climate change and growing global food demand.

%Bibliography
\bibliographystyle{plainnat}
\bibliography{paper}

\end{document}